\documentclass[11pt]{article}

\usepackage[final]{acl}

\usepackage{times}
\usepackage{latexsym}

\usepackage[T1]{fontenc}

\usepackage[utf8]{inputenc}

\usepackage{microtype}
\usepackage{hyperref}
\usepackage{url}
\usepackage{booktabs}
\usepackage{xspace}
\usepackage{graphicx}
\usepackage{textcomp}
\usepackage{multicol}
\usepackage{multirow}
\usepackage{enumitem}
\usepackage{pifont}
\usepackage{amsmath}
\usepackage{subfigure}
\usepackage{float}
\usepackage{tikz}
\usepackage{subcaption}
\usepackage{lineno}
\usepackage{amssymb}
\usepackage{bbm}
\usepackage{inconsolata}
\usepackage{placeins}

\usepackage{graphicx}

\definecolor{darkblue}{rgb}{0, 0, 0.5}
\hypersetup{colorlinks=true, citecolor=darkblue, linkcolor=darkblue, urlcolor=darkblue}

\def \name{\textsc{DieT}\xspace}
\definecolor{mycolor}{RGB}{33, 95, 154}

\title{Diet Your LLM: Dimension-wise Global Pruning of LLMs \\ via Merging Task-specific Importance Score}

\author{
  Jimyung Hong \quad
  Jaehyung Kim \\
  Yonsei University \\
  \texttt{\{jim0527,jaehyungk\}@yonsei.ac.kr}
}

\begin{document}
\maketitle

\begin{abstract}
Large language models (LLMs) have demonstrated remarkable capabilities, but their massive scale poses significant challenges for practical deployment. 
Structured pruning offers a promising solution by removing entire dimensions or layers, yet existing methods face critical trade-offs: task-agnostic approaches cannot adapt to task-specific requirements, while task-aware methods require costly training to learn task adaptability. 
We propose \name{} (\textbf{Di}mension-wise global pruning of LLMs via m\textbf{e}rging \textbf{T}ask-wise importance scores), a training-free structured pruning method that combines dimension-level granularity with task-aware selection. 
\name{} profiles activation magnitudes across tasks using only 100 samples per task, then applies majority voting to construct a single global mask.
Experiments on seven zero-shot benchmarks using Gemma-2 2B and 9B models demonstrate the effectiveness of \name{}. At 20\% sparsity on Gemma-2 2B, \name{} improves average accuracy by 2.1 percentage points over SliceGPT and by 9.2 percentage points over PuDDing. On Gemma-2 9B, \name{} remains competitive with SliceGPT while consistently outperforming PuDDing at 10\% and 20\% sparsity.\footnote{Code: \url{https://github.com/Jimmy145123/DIET}}
\end{abstract}
\section{Introduction}

Large Language Models (LLMs) have rapidly advanced with increasing scale, yet deploying them on resource-constrained platforms remains challenging due to strict compute and memory limits. 
This mismatch between model size and hardware budgets has motivated approaches to reduce their computational and memory footprints. 
Among them, pruning is one of the representative techniques, which removes model parameters or activations (\textit{i.e.}, masks them to zero to induce sparsity) while striving to preserve the original performance. 
Pruning has long been an active area of research in deep learning, and this interest has recently extended to LLMs. 
To make pruning practical for real-world deployment, inducing \textit{structured} sparsity patterns, rather than simply increasing the sparsity ratio, is crucial, since unstructured pruning is often incompatible with efficient computation using sparse matrix operations.


Prior works on structured pruning in LLMs can be categorized based on (1) the granularity of pruning (\textit{e.g.}, individual parameters, attention heads, or entire transformer blocks) and (2) whether the pruning decisions leverage task-specific information. 
For instance, SliceGPT~\citep{ashkboos2024slicegpt} exemplifies dimension-level, task-agnostic pruning by identifying less important dimensions through principal component analysis (PCA) of hidden embeddings using calibration data.
However, its task-agnostic mask cannot adapt to task-wise variation, and PCA-based identification introduces non-trivial computational overhead. 
PuDDing~\citep{wee2025pudding}, in contrast, adopts block-level, task-specific pruning by training a per-prompt router to dynamically skip selected Transformer blocks. 
While this approach achieves strong task-specific compression, it requires large training cost for the router and removes all attention and MLP computations within pruned blocks, limiting its generalization capability. 
These observations motivate us to explore an underexplored point in the design space: a \textit{dimension-level, task-aware, training-free} pruning framework for LLMs.

In this work, we propose \name{}, a framework for \textbf{Di}mension-wise global pruning of LLMs via m\textbf{e}rging \textbf{T}ask-wise importance scores. 
\name{} requires no additional training, such as model fine-tuning or router training, for pruning; it only relies on streamed model activations from a few task-specific samples to generate a simple, deployable global mask. 
To be specific, for each task, \name{} first profiles the outputs of MLP layers at every Transformer block and computes a per-dimension importance vector by averaging absolute activations. 
It then converts the lowest-ranked dimensions into a task-specific binary pruning mask and stacks these selectors across tasks. 
By aggregating masks across tasks, \name{} identifies dimensions with the highest cross-task agreement to achieve a target sparsity level, forming a single global pruning mask. 
This mask is uniformly applied to all residual-connected linear layers: input-dimension masking for embeddings, attention projections, MLP input projections, and the LM head and output-dimension masking for attention and MLP output projections, respectively.

We demonstrate the effectiveness of \name{} on Gemma-2 \citep{team2024gemma2} (2B and 9B) at 10\%, 20\%, and 30\% dimension sparsity under zero-shot evaluation on 7 different benchmarks.
Across both model sizes, \name{} consistently outperforms the
router-based pruning baseline at 10\% and 20\% sparsity, while
remaining competitive with SliceGPT.

\noindent Overall, our key contributions are as follows: 
\begin{itemize}[leftmargin=5.5mm,topsep=0pt]
    \vspace{0.05in}
    \item[$\circ$] We propose \name{}, a dimension-wise global pruning framework that yields a single global mask which is easy to apply. \vspace{-0.1in}
    \item[$\circ$] We introduce a simple activation-based profiling requiring no additional training. \vspace{-0.1in}
    \item[$\circ$] \name{} achieves comprehensive zero-shot results across two model scales and seven benchmarks, consistently outperforming PuDDing and remaining competitive with SliceGPT at 10\% and 20\% sparsity.
    \vspace{-0.1in} 
    \item[$\circ$] Empirically, \name{} exhibits an advantage over task-wise depth pruning algorithms, achieving improvements of more than 20 percentage points over the depth-pruning baseline on several commonsense reasoning tasks.
\end{itemize}

\begin{figure*}[ht]
\vspace{-0.1in}
\centering
\includegraphics[width=\linewidth]{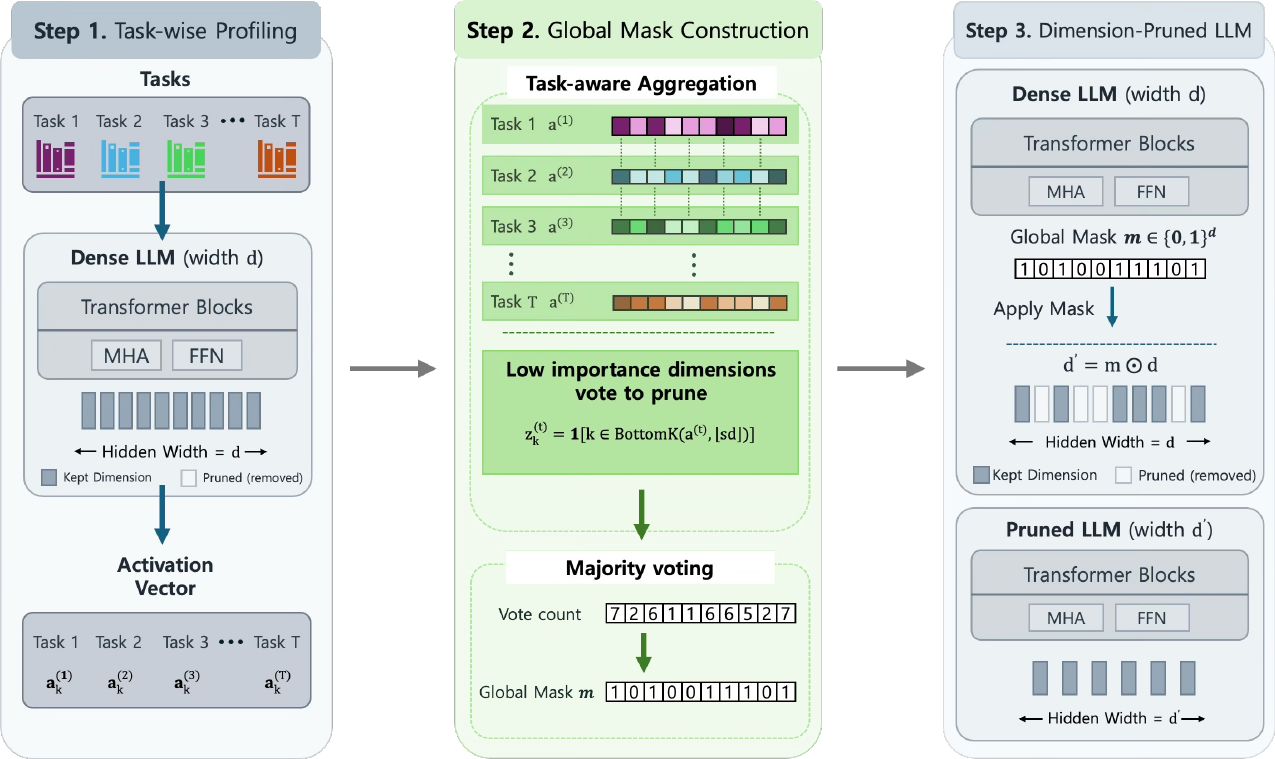}
\vspace{-0.25in}
\caption{
\textbf{Step 1.} Task-specific samples are used to profile a dense LLM and compute task-wise activation vectors. 
\textbf{Step 2.} Low-importance dimensions vote to prune, and the votes are merged by majority voting into a global mask $m \in \{0,1\}^d$. 
\textbf{Step 3.} The mask is applied across layers, producing a pruned LLM with reduced hidden width $d'$.
}
\vspace{-0.2in}
\label{fig:method}
\end{figure*}

\section{Dimension-wise Global Pruning via Merging Task-wise Importance Scores}
In this section, we introduce \name{}, a \textbf{Di}mension-wise structured pruning method for LLMs that merg\textbf{e}s \textbf{T}ask-wise importance signals to produce a single global pruning mask applicable across all tasks. 
The overview of \name{} is presented in Fig.~\ref{fig:method}.

\subsection{Preliminaries}
\label{sec:prelim}

Let $\mathcal{M}$ denote a transformer-based LLM with \(B\) transformer blocks.
Without loss of generalization, we assume the following with respect to $\mathcal{M}$:
\begin{itemize}[leftmargin=*,noitemsep,topsep=2pt]
    \item Each transformer block has an identical structure which consists of a multi-head attention layer followed by an MLP layer, both with residual connections and layernorms (pre- or post-).
    \item It maintains a fixed hidden dimension \(d\) throughout all blocks, which serves as the input/output dimension for most linear sublayers: word embeddings, attention projections, MLP layers, and LM head. 
    For example, Gemma-2~\citep{team2024gemma2} uses \(d=2304\) in 2B model and \(d=4608\) in 9B model across all \(B\) blocks.
\end{itemize}
Then, the goal of structured pruning is to remove entire computational units (\textit{e.g.}, dimensions, attention heads, or blocks), while unstructured pruning zeroes individual weights of sublayers without such consideration. 
In this work, we mainly consider dimension-wise pruning, because the above architectural property enables a single global mask over \(d\) to be applied uniformly to all layers, without requiring per-layer or per-task customization.

\paragraph{Mathematical formulation.} Let \(y = Wx + b\) denote a generic linear sublayer in each block, where \(W \in \mathbb{R}^{m \times n}\), \(b \in \mathbb{R}^m\), \(x \in \mathbb{R}^n\), and \(y \in \mathbb{R}^m\). 
Then, depending on the type of sublayer, the dimensions $m$ and $n$ could be different:
\begin{itemize}[leftmargin=*,noitemsep,topsep=2pt]
    \item \emph{Input} dimension \(n=d\): word embeddings, attention query/key/value projections, MLP input projection, and LM head.
    \item \emph{Output} dimension \(m=d\): attention output projection and MLP output projection.
\end{itemize}
Then, for each case, we consider different operations for dimension-wise pruning. 
For layers with input dimension \(n = d\), we apply \emph{input-dimension masking} using a binary selector \(\mathbf{z} \in \{0,1\}^{d}\):
\begin{equation}\label{eq:masking-input}
    y \;=\; W P_{\mathbf{z}}\, x + b,\qquad P_{\mathbf{z}}=\mathrm{diag}(\mathbf{z})
\end{equation}
which zeros the columns of \(W\) where \(z_j=0\).

For layers with output dimension \(m = d\), we apply \emph{output-dimension masking} using the same selector $\mathbf{z}$:
\begin{equation}\label{eq:masking-output}
    \tilde{y} \;=\; \tilde{P}_{\mathbf{z}}\, (W x + b),\qquad \tilde{P}_{\mathbf{z}}=\mathrm{diag}(\mathbf{z})
\end{equation}
which zeros the rows of \(W\) where \(z_j=0\). 
We refer to \eqref{eq:masking-input}–\eqref{eq:masking-output} collectively as the \emph{generic masking operator}.
The application of a masking operator reduces the effective dimension of $\mathcal{M}$ from $d$ to $d'=\sum_{j=1}^{d}z_j$.
Throughout this work, we apply the same global mask \(\mathbf{z} \in \{0,1\}^{d}\) to all parametrized sublayers (\textit{i.e.}, except normalization) via the appropriate operator (input or output masking depending on the layer's input/output dimensions).

\subsection{\name{}: Merge Task-wise Importance for Global Dimension-wise Pruning of LLMs}
\label{sec:diet-method}

\name{} derives a global pruning mask through a two-phase process. 
First, \name{} identifies the dimensions that contribute minimally to each task.
Then, \name{} aggregates these task-specific assessments: dimensions that most tasks agree to be uninformative are masked across all sublayers.

\paragraph{Task-wise importance scoring.}
\label{sec:task-scoring}
For each task \(t \in \mathcal{T}\), we assess dimension importance by measuring activation magnitudes using task-specific data. 
Activation magnitude serves as a widely-adopted proxy for feature contribution: dimensions with consistently low activations contribute little to the model's outputs and are candidates for removal~\citep{sun2024wanda,molchanov2016pruning}.

Specifically, we profile the pretrained LLM \(f_\theta\) on \(N\) samples \(\{x^{(t)}_i\}_{i=1}^{N}\) from task \(t\), recording activations at the output of the MLP layer in each transformer block (i.e., before the post-MLP layernorm and the residual add).
Let \(\mathcal{B}\) denote the set of transformer blocks (with \(B = |\mathcal{B}|\)). 
For sample \(x^{(t)}_i\), let \(h^{(t)}_{i,b} \in \mathbb{R}^{L_i \times d}\) represent the MLP outputs at block \(b\), where \(L_i\) is the number of valid (non-padding) tokens and columns index residual dimensions.
We compute a per-dimension importance score \(a^{(t)}_k\) by averaging absolute activations first across all blocks (capturing cross-layer consistency) and then across all tokens and samples:
\begin{equation}
\bar{h}^{(t)}_{i}[j,k] \;=\; \frac{1}{B}\sum_{b\in\mathcal{B}} \bigl|\,h^{(t)}_{i,b}[j,k]\,\bigr|,
\end{equation}
\begin{equation}
a^{(t)}_k \;=\;
{\displaystyle \sum_{i=1}^{N}\sum_{j=1}^{L_i} \bar{h}^{(t)}_{i}[j,k]}~\Big/~
     {\displaystyle \sum_{i=1}^{N} L_i}.
\end{equation}
The resulting vector \(\mathbf{a}^{(t)} \in \mathbb{R}_{\geq 0}^{d}\) quantifies each dimension's contribution to task \(t\). 
Given target sparsity \(s \in [0,1]\), we identify the \(\lfloor s \cdot d \rfloor\) dimensions with smallest \(a^{(t)}_k\) as pruning candidates, producing a binary selector \(\mathbf{z}^{(t)} \in \{0,1\}^{d}\) where \(z^{(t)}_k = 1\) indicates a pruning vote for dimension \(k\).

\paragraph{Merging into global pruning mask.}
\label{sec:merging}
Each task produces a pruning vote reflecting its own data distribution, but our objective is a single pruning decision applicable to all tasks. 
To this end, we employ a simple idea of \textit{majority voting}: dimensions that multiple tasks independently flag as unimportant are likely safe to remove globally without significantly harming individual tasks.

Concretely, we aggregate votes by stacking task-wise selectors into a matrix \(Z = [\mathbf{z}^{(t)}]_{t \in \mathcal{T}} \in \{0,1\}^{d \times T}\) and counting votes per dimension: \(c_k = \sum_{t \in \mathcal{T}} z^{(t)}_k\). 
The indices for global pruning \(\mathcal{O}\) consist of \(\lfloor s \cdot d \rfloor\) dimensions with highest vote counts:
\begin{equation}
    \mathcal{O} \;=\; \arg\text{top}_{\lfloor s \cdot d \rfloor}\,\bigl\{\,c_k\,\bigr\}_{k=1}^{d},
\end{equation}
where \(\arg\text{top}_k\) selects the indices of the \(k\) largest values. 
We define the omission mask \(\mathbf{m} \in \{0,1\}^{d}\) by \(m_k = \mathbbm{1}\{k \in \mathcal{O}\}\), and let \(\mathbf{u} = \mathbf{1} - \mathbf{m}\) denote the keep selector with \(P = \mathrm{diag}(\mathbf{u})\) in Eq.\ref{eq:masking-input} and \ref{eq:masking-output}.

We apply this global mask to all parametrized sublayers uniformly: 
for every layer with input dimension \(d\) (embeddings, attention query/key/value projections, MLP input projections, LM head), we apply input-dimension masking via \(W \leftarrow WP\) (Eq.~\ref{eq:masking-input}), which zeros the columns of \(W\) corresponding to dimensions in \(\mathcal{O}\).
For every layer with output dimension \(d\) (attention output projection, MLP output projection), we apply output-dimension masking via \(W \leftarrow \tilde{P}W\) (Eq.~\ref{eq:masking-output}), which zeros the corresponding rows.
This mask is applied identically across all \(B\) transformer blocks, yielding a single architectural modification: all sublayers now operate on the same pruned set of dimensions.
This modification is task-agnostic, \textit{i.e.}, the pruned model serves all tasks without per-task customization.

\subsection{From Global Masking to Hard Pruning for Computational Efficiency}
\label{sec:hard-pruning}

The global mask derived in Sec.~\ref{sec:diet-method} specifies which residual dimensions are removed across all sublayers.
This mask is applied via the generic masking operators in Eq.~\ref{eq:masking-input}–\ref{eq:masking-output}, \textit{i.e.}, zero masking, which is sufficient for evaluating the quality of dimension selection in terms of accuracy.

For efficiency evaluation, we additionally consider an \emph{optional hard pruning} setting, where masked dimensions are physically removed and the hidden dimension is compacted from \(d\) to
\begin{equation}
d' \;=\; \sum_{j=1}^{d} u_j,
\end{equation}
with \(\mathbf{u} = \mathbf{1} - \mathbf{m}\) denoting the keep selector in Sec.~\ref{sec:merging}.
This compaction is applied uniformly across all residual-connected sublayers, resulting in a reduced hidden dimension.
However, this hard pruning changes the input dimensionality of subsequent linear sublayers; 
under RMSNorm or LayerNorm, assuming independent and equally-scaled input features, reducing the input dimension from \(d\) to \(d'\) decreases the expected activation variance by a factor of \(d'/d\).
Without correction, this variance reduction can lead to degraded representations.

\paragraph{Variance correction via weight rescaling.}
To compensate for this effect, we rescale the weights of linear sublayers whose \emph{input dimension} is reduced.
Specifically, we apply the scaling factor
\begin{equation}
\alpha \;=\; \sqrt{\frac{d}{d'}}.
\end{equation}
For a linear sublayer whose input dimension is sliced to \(d'\), the corresponding weight matrix is rescaled as
\begin{equation}
W_{\text{scaled}} \;=\; \alpha \cdot W_{\text{sliced}},
\end{equation}
where \(W_{\text{sliced}}\) denotes the weight matrix after removing masked input dimensions. 
For sublayers where only the output dimension is reduced (\textit{e.g.}, attention output projection or MLP down-projection), the input dimensions remain unchanged and no rescaling is applied.
Normalization parameters are sliced to the retained dimensions without scaling.

Unless stated otherwise, all accuracy-focused results are obtained
using zero masking, while hard pruning with variance correction is
used for the efficiency evaluation and the dedicated comparison
with zero masking in Table~\ref{tab:hard_pruning}.

\section{Experiments}\label{sec:4}


\begin{table*}[ht!]
\caption{Zero-shot accuracy comparisons of \name{} against baseline pruning algorithms on Gemma-2 2B on seven NLP benchmarks. The best performances are marked in \textbf{bold}, and the runner-up is marked with \underline{underline}. Tasks with * are presented with \textit{acc\_norm} and remaining tasks with \textit{acc}.}
\label{tab:main_tab_2b}
\centering
\resizebox{0.95\textwidth}{!}{%
\begin{tabular}{@{}lccccccccc@{}} 
\toprule
\multirow{2}{*}{Method} & \multirow{2}{*}{\begin{tabular}[c]{@{}c@{}}Sparsity\end{tabular}} &
\multicolumn{7}{c}{Per-task Accuracies (\%)} &
\multirow{2}{*}{\begin{tabular}[c]{@{}l@{}}Average\\ Acc. (\%)\end{tabular}} \\ \cmidrule(lr){3-9}
 &  & \multicolumn{1}{l}{BoolQ} & RTE & HellaSwag* & WinoGrande & Arc-E* & Arc-C* & OBQA* & \\ \midrule

No pruning (Original) & 0 & 73.5 & 61.4 & 73.0 & 68.4 & 80.2 & 49.7 & 41.8 & 64.0 \\ \midrule

Magnitude-Dim & 20\% & 38.2 & \textbf{55.2} & 31.1 & 50.0 & 39.8 & 21.3 & 25.0 & 37.2 \\
SliceGPT                   & 20\% & 38.4 & 52.7 & \textbf{43.5} & \textbf{60.9} & \underline{47.0} & \textbf{27.3} & \textbf{30.6} & \underline{42.9} \\
PuDDing                    & 20\% & \underline{43.4} & 53.8 & 25.8 & 49.8 & 27.3 & 23.6 & 27.0 & 35.8 \\
\midrule
\name{} (Ours)                & 20\% & \textbf{63.3} & \underline{54.9} & \underline{41.8} & \underline{53.2} & \textbf{49.8} & \underline{24.1} & \underline{27.6} & \textbf{45.0} \\
\midrule

Magnitude-Dim & 10\% & 51.3 & 52.7 & 44.8 & 53.9 & 59.6 & 32.5 & 31.2 & 46.6 \\
SliceGPT                   & 10\% & \underline{68.1} & \underline{54.2} & \underline{60.0} & \textbf{66.2} & \underline{66.4} & \underline{38.3} & \textbf{35.8} & \underline{55.6} \\
PuDDing                    & 10\% & 42.6 & 51.3 & 25.8 & 50.8 & 27.3 & 23.7 & 30.0 & 35.9 \\
\midrule
\name{} (Ours)                & 10\% & \textbf{70.6} & \textbf{54.9} & \textbf{62.7} & \underline{62.8} & \textbf{69.0} & \textbf{38.4} & \underline{35.4} & \textbf{56.3} \\


\bottomrule
\end{tabular}%
}
\end{table*}

\subsection{Setups}\label{sec:4.1}

\paragraph{Datasets.}
\label{sec:Datasets.}
We evaluate \name{} on seven widely-used NLP benchmarks covering diverse reasoning and language understanding capabilities:
{(1) \textit{BoolQ}}~\citep{clark2019boolq} contains yes/no questions derived from naturally occurring queries paired with Wikipedia passages.
{(2) \textit{RTE}}~\citep{dagan2005pascal} (Recognizing Textual Entailment) evaluates whether a hypothesis is entailed by a given premise.
{(3) \textit{HellaSwag}}~\citep{zellers2019hellaswag} tests commonsense reasoning through sentence completion requiring contextual understanding.
{(4) \textit{WinoGrande}}~\citep{sakaguchi2021winogrande} measures commonsense reasoning via pronoun disambiguation.
{(5, 6) \textit{ARC-Challenge and ARC-Easy}}~\citep{clark2018arc} provide grade-school science questions at varying difficulty levels.
{(7) \textit{OpenBookQA}}~\citep{mihaylov2018openbookqa} requires multi-hop reasoning over elementary science facts.

For task-wise importance scoring (Sec.~\ref{sec:task-scoring}), we randomly sample 100 examples per task from the training split of each dataset.
Final evaluations use the native splits defined by the lm-evaluation-harness~\citep{eval-harness}: BoolQ, RTE, HellaSwag, and WinoGrande are evaluated on their validation splits, while ARC-Easy, ARC-Challenge, and OpenBookQA use test splits.  

\paragraph{Metrics.} 
\label{sec:Metrics.}
We evaluate with two metrics: standard accuracy (\textit{acc}) and length-normalized accuracy (\textit{acc\_norm}). For multiple-choice datasets with variable-length options—HellaSwag, ARC-Easy, ARC-Challenge, and OpenBookQA—we report \textit{acc\_norm} as the primary metric because it mitigates length bias by dividing each option’s log-likelihood by its token length. For BoolQ, RTE, and WinoGrande, we follow the default \texttt{lm-eval-harness} task definitions and report \textit{acc} only, since answers are short (often single-token verbalizers), making length normalization unhelpful and noisy. We present per-task scores and an unweighted average across all seven tasks.

\paragraph{Baselines.}
\label{sec:Baselines.}
We compare \name{} with three structured pruning baselines under matched sparsity ratios. 
\textsc{(1) Magnitude–Dim}~\citep{han2015learning}: dimension-wise pruning that aggregates absolute-weight scores per residual dimension and applies a single global dimension mask uniformly to all residual-connected linear layers, zeroing corresponding rows or columns. 
\textsc{(2) SliceGPT}~\citep{ashkboos2024slicegpt}: rotate-and-slice compression that learns an orthogonal activation basis from a calibration corpus and slices MLP and attention projections to a reduced hidden width implied by the target sparsity.
\textsc{(3) PuDDing}~\citep{wee2025pudding}: prompt-conditioned depth pruning in which trained lightweight router selects, per input, which transformer blocks to skip. 
By omitting layers selected by trained router, it achieves target sparsity ratio. 


\begin{table*}[ht!]
\caption{Zero-shot accuracy comparisons of \name{} against baseline pruning algorithms on Gemma-2 9B on seven NLP benchmarks. The best performances are marked in \textbf{bold}, and the runner-up is marked with \underline{underline}. Tasks with * are presented with \textit{acc\_norm} and remaining tasks with \textit{acc}.}
\label{tab:main_tab_9b}
\centering
\resizebox{0.95\textwidth}{!}{%
\begin{tabular}{@{}lccccccccc@{}} 
\toprule
\multirow{2}{*}{Method} & \multirow{2}{*}{\begin{tabular}[c]{@{}c@{}}Sparsity\end{tabular}} &
\multicolumn{7}{c}{Per-task Accuracies (\%)} &
\multirow{2}{*}{\begin{tabular}[c]{@{}l@{}}Average\\ Acc. (\%)\end{tabular}} \\ \cmidrule(lr){3-9}
 &  & \multicolumn{1}{l}{BoolQ} & RTE & HellaSwag* & WinoGrande & Arc-E* & Arc-C* & OBQA* & \\ \midrule

No pruning (Original) & 0 & 84.4 & 67.9 & 80.0 & 74.2 & 87.8 & 66.1 & 47.2 & 72.5 \\ \midrule

Magnitude-Dim & 20\% & 68.5 & 53.8 & 49.5 & 60.6 & 62.0 & 30.8 & 31.6 & 51.0 \\
SliceGPT                   & 20\% & 49.2 & 54.2 & 54.5 & 68.3 & 62.4 & 36.1 & 39.4 & \textbf{52.0} \\
PuDDing                    & 20\% & 62.8 & 57.4 & 49.8 & 59.7 & 51.3 & 32.1 & 29.2 & 48.9 \\
\midrule
DIET (Ours)                & 20\% & 74.7 & 54.2 & 48.5 & 59.2 & 61.7 & 32.0 & 31.2 & \underline{51.6} \\
\midrule

Magnitude-Dim & 10\% & 82.7 & 62.1 & 71.1 & 71.4 & 80.2 & 50.6 & 43.2 & \textbf{65.9} \\
SliceGPT                   & 10\% & 77.6 & 56.3 & 69.7 & 74.0 & 78.7 & 51.0 & 45.2 & \underline{64.7} \\
PuDDing                    & 10\% & 64.5 & 64.3 & 49.5 & 59.5 & 49.1 & 30.3 & 30.2 & 49.7 \\
\midrule
DIET (Ours)                & 10\% & 82.8 & 62.1 & 69.9 & 68.7 & 79.1 & 49.6 & 40.0 & 64.6 \\


\bottomrule
\end{tabular}%
}
\end{table*}

\paragraph{Implementation details.}
\label{sec:Implementation.}
In our experiments, we employ \texttt{Gemma-2} models \citep{team2024gemma2} with different sizes of 2B and 9B. 
For the \texttt{Task-wise importance scoring} phase, we randomly sampled 100 samples from each task to accelerate the activation profiling process (the effect of sample size on zero shot performance is analyzed in Figure \ref{fig:sample-number}). 
To ensure reproducibility and mitigate the influence of random sampling, we set the random seed to 42 throughout all experiments.
For baseline methods, we follow the default configurations from their released implementations. SliceGPT~\citep{ashkboos2024slicegpt} uses WikiText2~\citep{merity2016wikitext2} as the calibration corpus with 64 samples and a maximum sequence length of 2048. 
To ensure fair comparison with \name{}, we also evaluate SliceGPT using 100 samples per task from the seven zero-shot benchmarks described in Section~\ref{sec:Datasets.}.
PuDDing~\citep{wee2025pudding} trains a BERT-base~\citep{devlin2019bert} router for 10 epochs using AdamW~\citep{loshchilov2017adamw} optimizer with an 80/20 train/test split. 
We trained router on the same benchmark suite, mentioned in \ref{sec:Datasets.}. 
Magnitude-based methods apply global L1 pruning, with the structured variant following \eqref{eq:masking-input}–\eqref{eq:masking-output} to mask dimensions uniformly across residual-connected layers.
All methods are evaluated at 10\%, 20\%, and 30\% sparsity ratios and all evaluations use lm-evaluation-harness~\citep{eval-harness} in zero-shot mode with fixed seed 42.  

\subsection{Main results}
Table \ref{tab:main_tab_2b} presents zero-shot accuracy on seven benchmarks for Gemma-2 2B~\citep{team2024gemma2}, comparing \name{} against structured pruning SliceGPT~\citep{ashkboos2024slicegpt} and PuDDing~\citep{wee2025pudding}, as well as magnitude-based methods~\citep{han2015learning}. 
Here, \name{} consistently outperforms both structured pruning methods across all sparsity levels. 
For example, at 10\% sparsity, \name{} achieves 56.3\% average accuracy, exceeding SliceGPT by 0.7 percentage points and PuDDing by 20.4 percentage points.
This gap persists at 20\% sparsity, where \name{} (45.0\%) surpasses SliceGPT (42.9\%) by 2.1 percentage points and PuDDing (35.8\%) by 9.2 percentage points. 
Notably, while SliceGPT and PuDDing exhibit performance degradation, \name{} preserves more task-specific capabilities on average.
Per-task analysis reveals this advantage holds broadly: at 10\% sparsity, \name{} outperforms SliceGPT on five of the seven tasks and PuDDing on all seven tasks, with particularly strong results on BoolQ (70.6\%), HellaSwag (62.7\%), and ARC-Easy (69.0\%). 
At 20\%, \name{} exceeds SliceGPT on three tasks and PuDDing on all seven tasks, demonstrating robustness on reasoning tasks like BoolQ and ARC-Easy.

\begin{table}[t]
\caption{
Memory and computational efficiency comparison of the original,
zero-masked, and hard-pruned \name{} models on Gemma-2 2B at
20\% sparsity.
}
\label{tab:efficiency_2b}
\centering
\resizebox{\columnwidth}{!}{%
\setlength{\tabcolsep}{5pt}
\renewcommand{\arraystretch}{1.1}
\begin{tabular}{lcccc}
\toprule
Model
& \begin{tabular}[c]{@{}c@{}}Avg. Task-wise\\Peak Memory (MB)\end{tabular}
& \begin{tabular}[c]{@{}c@{}}Memory\\Reduction\end{tabular}
& \begin{tabular}[c]{@{}c@{}}Estimated FLOPs\\per Batch ($10^{12}$)\end{tabular}
& \begin{tabular}[c]{@{}c@{}}FLOPs\\Reduction\end{tabular} \\
\midrule
Original
& 7244.7
& ---
& 4.62
& --- \\

\name{} (Zero-mask)
& 7244.7
& $0.0\%$
& 4.62
& $0.0\%$ \\

\name{}-Hard
& \textbf{6247.2}
& \textbf{$13.8\%$}
& \textbf{3.70}
& \textbf{$20.0\%$} \\

\bottomrule
\end{tabular}%
}
\end{table}
\begin{table*}[ht!]
\caption{Zero-shot accuracy comparisons of the standard zero-masking \name{} and the hard-pruning implementation (\name{}-Hard) on Gemma-2 2B. \name{}-Hard physically removes dimensions and applies variance rescaling to mitigate distribution shift. The better performance between the two pruned methods is marked in \textbf{bold}. Tasks with * are presented with \textit{acc\_norm} and remaining tasks with \textit{acc}.}
\label{tab:hard_pruning}
\centering
\resizebox{0.95\textwidth}{!}{%
\begin{tabular}{@{}lccccccccc@{}}
\toprule
\multirow{2}{*}{Method} & \multirow{2}{*}{Sparsity} &
\multicolumn{7}{c}{Per-task Accuracies (\%)} &
\multirow{2}{*}{\begin{tabular}[c]{@{}l@{}}Average\\ Acc. (\%)\end{tabular}} \\ 
\cmidrule(lr){3-9}
 &  & BoolQ & RTE & HellaSwag* & WinoGrande & Arc-E* & Arc-C* & OBQA* & \\ 
\midrule

No pruning (Original) & 0 & 73.5 & 61.4 & 73.0 & 68.4 & 80.2 & 49.7 & 41.8 & 64.0 \\ 
\midrule

\name{} (Zero-mask) & 20\% & 63.3 & 54.9 & 41.8 & 53.2 & 49.8 & 24.1 & 27.6 & 45.0 \\
\name{} (Hard-prune) & 20\% & 64.7 & 51.6 & 42.5 & 54.7 & 51.3 & 25.9 & 27.4 & \textbf{45.4} \\

\bottomrule
\end{tabular}%
}
\end{table*}

Next, we additionally conduct the experiments with larger model.
Table \ref{tab:main_tab_9b} presents results on Gemma-2 9B model at 10\% and 20\% sparsity levels. 
Similar to the results on 2B model, \name{} remains competitive with structured pruning baselines across model scales.
For instance, at 10\% sparsity, \name{} achieves 64.6\% average accuracy, trailing SliceGPT (64.7\%) by only 0.1 percentage points while substantially outperforming PuDDing (49.7\%) by 14.9 percentage points. 
At 20\% sparsity, \name{} (51.6\%) remains competitive with SliceGPT (52.0\%) and surpasses PuDDing (48.9\%) by 2.7 percentage points.

The comparison across model scales reveals consistent patterns. 
\name{} remains competitive with SliceGPT on both 2B and 9B models, with the absolute gap remaining within 2.1 percentage points across all evaluated settings.
Against PuDDing, \name{} demonstrates stronger improvements on the smaller 2B model (9.2 percentage points) compared to 9B (2.7 percentage points), suggesting that the advantage of task-aware dimension selection over prompt-conditioned depth pruning narrows as model capacity grows.
While Magnitude-Dim achieves 51.0\% on 9B at 20\% sparsity—slightly below \name{}'s 51.6\%—\name{} demonstrates competitive performance among dimension-level structured approaches.
This competitive performance across two model scales spanning nearly $4\times$ in parameter count positions \name{} as a robust structured pruning method across different model scales.

\paragraph{Memory and computational efficiency.}
We report memory and computational efficiency results in terms of peak GPU memory usage and estimated FLOPs in Table~\ref{tab:efficiency_2b}. 
All efficiency measurements are conducted on Gemma-2 2B at a sparsity ratio of 20\% under the same forward-inference setting. 
The standard zero-masking implementation of \name{} retains the original dense tensor shapes and parameter count, resulting in no reduction in either peak GPU memory or estimated FLOPs. 
In contrast, \name{}-Hard physically removes the selected dimensions and compacts the hidden size, reducing average task-wise peak GPU memory from 7244.7\,MB to 6247.2\,MB and estimated inference FLOPs from $4.62\times10^{12}$ to $3.70\times10^{12}$ per batch. 
These correspond to reductions of 13.8\% and 20.0\%, respectively, indicating practical efficiency gains from structural hard pruning rather than zero masking. 

In addition to these efficiency improvements, we verify that hard pruning preserves predictive performance.
Table~\ref{tab:hard_pruning} compares the zero-shot accuracy of the standard zero-masking implementation of \name{} (reported in Table~\ref{tab:main_tab_2b}) with its hard-pruned counterpart under the same sparsity ratio. 
Despite physically removing dimensions and compacting the hidden size, \name{}-Hard maintains accuracy comparable to the zero-masked model across tasks. 
In particular, the average accuracy of \name{}-Hard (45.4\%) closely matches that of the zero-masked \name{} (45.0\%), demonstrating that the proposed variance correction effectively mitigates the distribution shift induced by dimensionality reduction.
These results indicate that hard pruning serves as a faithful and practically efficient realization of \name{} without sacrificing zero-shot performance.

\subsection{More Analyses}
\label{sec:ablation}

In this section, we present additional analyses of \name{} on the seven zero-shot benchmarks using Gemma-2 2B~\cite{team2024gemma2} unless otherwise noted. Due to space constraints, we report the main findings here and defer the remaining experiments to Appendix ~\ref{supp:analyses}. These include ablation studies on two mask-construction choices: the use of a merged global mask instead of single-task masks, and binary vote-based merging instead of continuous score aggregation. Together, these results validate that aggregating task-wise pruning signals into a single global mask is a robust design choice. 

\begin{figure}[t]
  \centering
  \includegraphics[width=\linewidth]{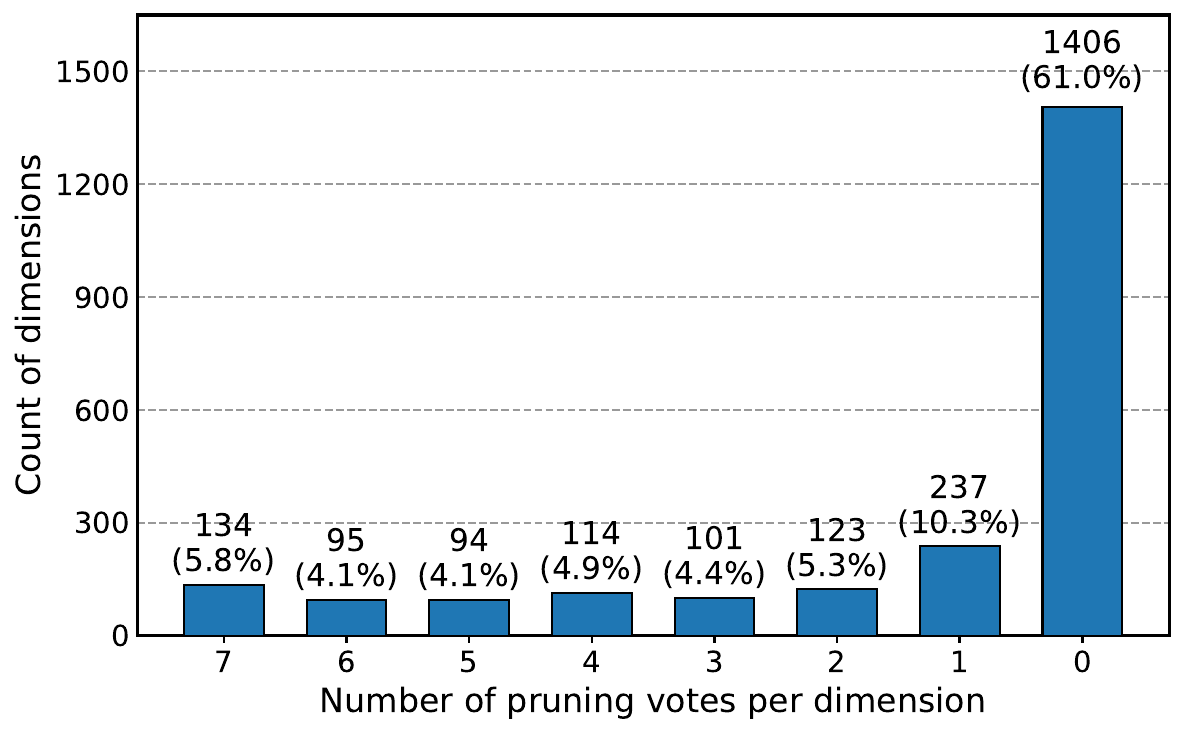} 
  \caption{Consensus-vote histogram by dimension. Numbers show total counts (and share of $d{=}2304$).}
  \label{fig:dimension-count}
  \vspace{-0.1in}
\end{figure}

\paragraph{Cross-task agreement via voting.}
We evaluate cross-task consensus on uninformative residual dimensions. For each dimension k, we compute total pruning votes $c_k{=}\sum_{t} z_k^{(t)}$ across the seven benchmarks (Figure~\ref{fig:dimension-count}). 
The histogram reveals striking consensus: dimensions flagged by nearly
all tasks (7 or 6 votes) account for $\approx$10\% of the hidden width
$d$, naturally justifying a 10\% sparsity target.
Expanding this threshold to $\geq 4$ votes encompasses
$\approx$20\% of dimensions, matching our target sparsity with strong
agreement.
This heavy vote concentration demonstrates that \name{} uncovers
universal architectural redundancies across diverse tasks, validating
our core assumption that multi-task consensus robustly guides
training-free pruning.

\begin{figure}[t]
  \centering
  \includegraphics[width=\linewidth]{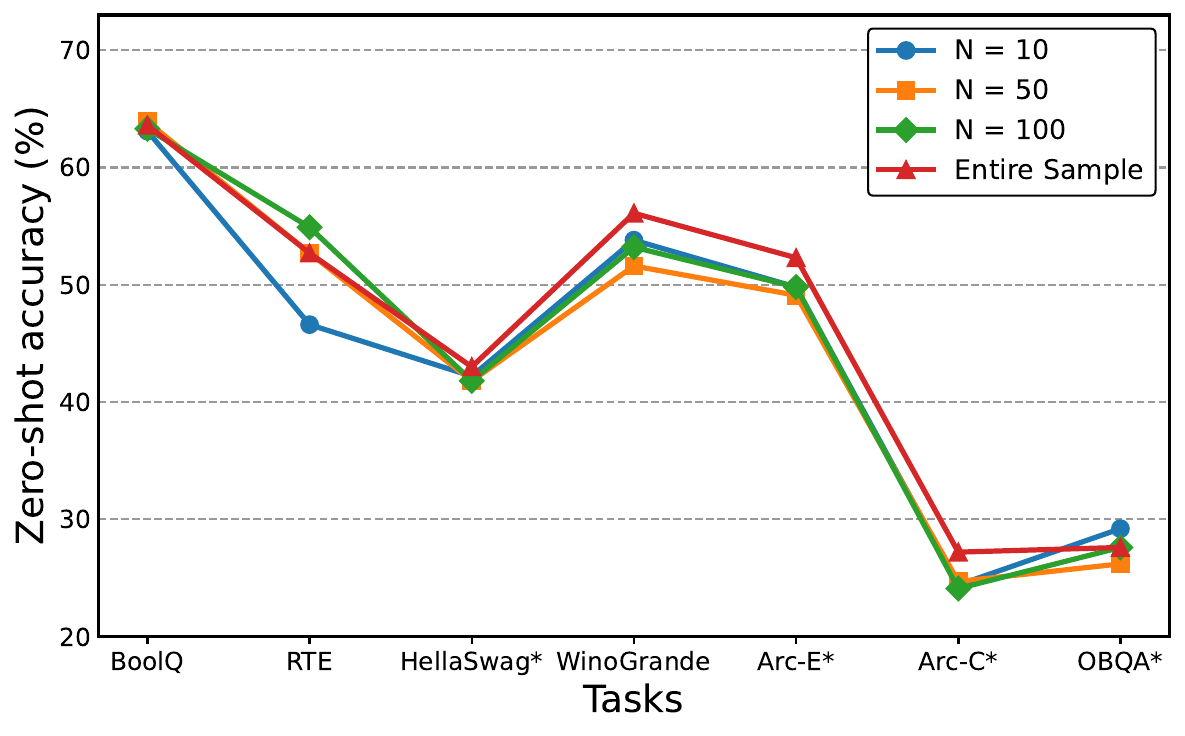} 
  \caption{Zero-shot accuracy by activation-profiling sample size. Benchmarks with * use \textit{acc\_norm}.}
  \label{fig:sample-number}
  \vspace{-0.1in}
\end{figure}

\paragraph{Robustness to profiling sample size.}
We investigate the activation-profiling budget by varying samples per task $N \in \{10, 50, 100, \text{full}\}$ under 20\% sparsity. As shown in Figure~\ref{fig:sample-number}, zero-shot accuracy improves consistently with $N$, rising from 44.2\% ($N=10$) to 46.1\% (full split). Crucially, the performance gap between a modest budget of $N=100$ (45.0\%) and the full baseline is only 1.1\%. This marginal difference indicates that \name{} reaches near-saturation with minimal calibration data, effectively balancing efficiency and downstream performance.

\paragraph{Generalization of \name{} to various LLMs.} 
\begin{table}[t]
\caption{Zero-shot average accuracy (\%) across seven benchmarks on Qwen2.5-7B, Phi-4-Mini-Reasoning, and Qwen3-14B-Base at 10\% and 20\% sparsity.
}
\label{tab:model_generality}
\resizebox{\columnwidth}{!}{%
\begin{tabular}{@{}lcccc@{}}
\toprule
Method & Sparsity & Qwen2.5-7B & Phi-4-mini & Qwen3-14B-Base \\ \midrule
No pruning & 0\% & 70.5 & 58.8 & 72.7 \\ \midrule
Magnitude - Dim & 10\% & 38.6 & 36.1 & 34.7 \\
\name{} (Ours) & 10\% & 60.9 & 46.3 & 66.7 \\ \midrule
\name{} (Ours) & 20\% & 37.0 & 37.2 & 49.0 \\
\bottomrule
\end{tabular}%
}
\vspace{-0.1in}
\end{table}
To verify cross-architecture generalizability, we evaluate \name{} on
Qwen2.5-7B~\citep{qwen2025qwen25technicalreport},
Phi-4-Mini-Reasoning (3.8B)~\citep{xu2025phi4mini_reasoning},
and Qwen3-14B-Base~\citep{yang2025qwen3technicalreport}
(Table~\ref{tab:model_generality}).
At 10\% sparsity, \name{} successfully preserves most original
performance across all three backbones without retraining.
However, at 20\% sparsity, performance drops more noticeably,
with absolute decreases of approximately 33.5, 21.6, and
23.7 percentage points on Qwen2.5-7B, Phi-4-Mini-Reasoning,
and Qwen3-14B-Base, respectively. This suggests that while activation-driven aggregation transfers robustly under light compression, aggressive pruning amplifies architecture-specific sensitivities, highlighting the need for adaptive layer-wise scaling mechanisms in future work.
\section{Related Works}
\label{sec:related}

\paragraph{Structured pruning of LLMs.}
Structured pruning methods are commonly categorized by pruning granularity into dimension-wise and block-wise approaches~\cite{ashkboos2024slicegpt,song2024sleb}.
Dimension-wise methods reduce layer width by removing less informative channels or embeddings; SliceGPT~\cite{ashkboos2024slicegpt} uses PCA-based alignment, while LLM-Pruner~\cite{ma2023llmpruner} and LoRAShear~\cite{chen2023lorashear} rely on gradient-based estimation or progressive structural optimization with recovery fine-tuning.
Block-wise methods instead reduce depth by removing redundant Transformer blocks using perplexity drops or iterative selection metrics~\cite{song2024sleb,kim2024shortenedllama}.

\paragraph{Task-aware pruning.}
Recent works explore dynamic, task-aware pruning to adapt compression to downstream tasks or inputs.
At the block level, PuDDing~\cite{wee2025pudding} trains a lightweight router to predict per-prompt block omission sets.
At the dimension level, Instruction-Following Pruning (IFPruning)~\cite{hou2025instruction} co-trains a dynamic sparsity predictor with the backbone LLM to generate input-dependent dimension masks.
Although these methods achieve high task adaptability, they require extensive training and introduce non-trivial runtime routing overhead.

\paragraph{Training-free pruning.}
Training-free approaches avoid the costs of router training, continued pre-training, or post-pruning recovery~\cite{ma2023llmpruner,chen2023lorashear,wee2025pudding,hou2025instruction}, making them practical for one-shot deployment.
Existing methods such as SLEB~\cite{song2024sleb}, Shortened-LLaMA~\cite{kim2024shortenedllama}, and SliceGPT~\cite{ashkboos2024slicegpt} compute fixed masks from general calibration corpora, making them task-agnostic and suboptimal for downstream demands.
In contrast, \name{} is both task-aware and training-free.
By profiling activation magnitudes on a small budget of task-specific samples and aggregating them via majority voting, \name{} yields a single global mask that captures multi-task signals without parameter updates or runtime routing.
\section{Conclusion}
We presented \name{}, a training-free dimension-wise global pruning
framework that merges task-wise importance estimates into a single
global mask.
By profiling MLP activations on per-task samples and aggregating
through majority voting, \name{} removes residual
dimensions that multiple tasks deem uninformative.
Across seven zero-shot benchmarks on Gemma-2 2B and 9B, \name{}
consistently outperforms PuDDing at 10\% and 20\% sparsity while
remaining competitive with SliceGPT, achieving an improvement of up
to 20.4 percentage points over PuDDing at 10\% sparsity.
In addition, \name{}-Hard reduces average task-wise peak GPU memory
by 13.8\% and estimated inference FLOPs by 20.0\%, while preserving
zero-shot performance comparable to the zero-masked implementation.
These results indicate that a router-free, activation-driven
approach can deliver task-aware compression with modest
profiling costs, positioning \name{} as a practical choice for
structured LLM pruning.

\newpage

\section*{Limitation and Future Work} 
Our primary evaluation focuses on seven English-centric zero-shot
benchmarks, with additional analyses on XNLI, IFEval, and MMLU.
However, multilingual coverage remains limited to a single NLI
benchmark, and broader evaluation across multilingual generation,
safety-critical tasks, long-context scenarios, and tool-use
applications remains unexplored.
At high sparsity levels on non-Gemma architectures, we observe notable
accuracy degradation, suggesting the need for adaptive sparsity
schedules, per-layer allocation strategies, or lightweight
post-pruning recovery techniques.
Also, task-weighted or distribution-aware voting mechanisms may
improve robustness, while smarter calibration policies, such as
difficulty-aware or uncertainty-driven sampling, could enhance
sample efficiency.

\section*{Broader Impact and Ethical Implications}

By reducing computational and memory requirements through a single global mask, \name{} offers practical benefits for model deployment and environmental sustainability. Unlike task-specific pruning methods that require maintaining multiple pruned variants, \name{}'s global mask produces a single task-aware model capable of handling diverse tasks effectively. This consolidation simplifies deployment pipelines, reduces storage overhead, and lowers both energy consumption during inference and the carbon footprint associated with model distribution. By eliminating the need for multiple model variants, \name{} makes large language models more accessible to resource-constrained settings while maintaining task-aware capabilities.
However, pruning alters internal representations and may shift model behavior in fairness, safety, and robustness. Practitioners should audit pruned models for bias and distribution shift, re-run safety evaluations, and transparently document sparsity levels and calibration procedures. We recommend using publicly available calibration data to respect privacy and pairing deployments with appropriate monitoring and safeguards.

\section*{Acknowledgments}

Jimyung and Jaehyung are affiliated with the Department of Artificial Intelligence at Yonsei University.
This research was supported in part by Institute for Information \& communications Technology Planning \& Evaluation (IITP) grant funded by the Korea government (MSIT) (No. RS-2020-II201361, Artificial Intelligence Graduate School Program (Yonsei University); No. RS-2025-25442405, Development of a Self-Learning World Model-Based
AGI System for Hyperspectral Imaging).
\bibliography{anthology, custom}

@inproceedings{ma2023llmpruner,
  title={Llm-pruner: On the structural pruning of large language models},
  author={Ma, Xinyin and Fang, Gongfan and Wang, Xinchao},
  journal={Advances in neural information processing systems},
  volume={36},
  pages={21702--21720},
  year={2023}
}

@article{chen2023lorashear,
  title={Lorashear: Efficient large language model structured pruning and knowledge recovery},
  author={Chen, Tianyi and Ding, Tianyu and Yadav, Badal and Zharkov, Ilya and Liang, Luming},
  journal={arXiv preprint arXiv:2310.18356},
  year={2023}
}

@inproceedings{ashkboos2024slicegpt,
  title={Slicegpt: Compress large language models by deleting rows and columns},
  author={Ashkboos, Saleh and Croci, Maximilian and Gennari do Nascimento, Marcelo and Hoefler, Torsten and Hensman, James},
  booktitle={International Conference on Learning Representations},
  volume={2024},
  pages={11682--11701},
  year={2024}
}

@inproceedings{wee2025pudding,
  title={Prompt-based Depth Pruning of Large Language Models},
  author={Wee, Juyun and Park, Minjae and Lee, Jaeho},
  journal={arXiv preprint arXiv:2502.04348},
  year={2025}
}

@inproceedings{sun2024wanda,
  title={A simple and effective pruning approach for large language models},
  author={Sun, Mingjie and Liu, Zhuang and Bair, Anna and Kolter, Zico},
  booktitle={International Conference on Learning Representations},
  volume={2024},
  pages={4942--4964},
  year={2024}
}

@inproceedings{molchanov2016pruning,
  title={Pruning convolutional neural networks for resource efficient inference},
  author={Molchanov, Pavlo and Tyree, Stephen and Karras, Tero and Aila, Timo and Kautz, Jan},
  journal={arXiv preprint arXiv:1611.06440},
  year={2016}
}

@inproceedings{hou2025instruction,
  title={Instruction-following pruning for large language models},
  author={Hou, Bairu and Chen, Qibin and Wang, Jianyu and Yin, Guoli and Wang, Chong and Du, Nan and Pang, Ruoming and Chang, Shiyu and Lei, Tao},
  journal={arXiv preprint arXiv:2501.02086},
  year={2025}
}

@article{kim2024shortenedllama,
  title={Shortened llama: Depth pruning for large language models with comparison of retraining methods},
  author={Kim, Bo-Kyeong and Kim, Geonmin and Kim, Tae-Ho and Castells, Thibault and Choi, Shinkook and Shin, Junho and Song, Hyoung-Kyu},
  journal={arXiv preprint arXiv:2402.02834},
  year={2024}
}

@inproceedings{song2024sleb,
  title={Sleb: Streamlining llms through redundancy verification and elimination of transformer blocks},
  author={Song, Jiwon and Oh, Kyungseok and Kim, Taesu and Kim, Hyungjun and Kim, Yulhwa and Kim, Jae-Joon},
  journal={arXiv preprint arXiv:2402.09025},
  year={2024}
}

@article{team2024gemma2,
  title={Gemma 2: Improving open language models at a practical size},
  author={Team, Gemma and Riviere, Morgane and Pathak, Shreya and Sessa, Pier Giuseppe and Hardin, Cassidy and Bhupatiraju, Surya and Hussenot, L{\'e}onard and Mesnard, Thomas and Shahriari, Bobak and Ram{\'e}, Alexandre and others},
  journal={arXiv preprint arXiv:2408.00118},
  year={2024}
}

@article{xu2025phi4mini_reasoning,
  title={Phi-4-mini-reasoning: Exploring the limits of small reasoning language models in math},
  author={Xu, Haoran and Peng, Baolin and Awadalla, Hany and Chen, Dongdong and Chen, Yen-Chun and Gao, Mei and Kim, Young Jin and Li, Yunsheng and Ren, Liliang and Shen, Yelong and others},
  journal={arXiv preprint arXiv:2504.21233},
  year={2025}
}

@misc{qwen2025qwen25technicalreport,
      title={Qwen2.5 Technical Report}, 
      author={Qwen and : and An Yang and Baosong Yang and Beichen Zhang and Binyuan Hui and Bo Zheng and Bowen Yu and Chengyuan Li and Dayiheng Liu and Fei Huang and Haoran Wei and Huan Lin and Jian Yang and Jianhong Tu and Jianwei Zhang and Jianxin Yang and Jiaxi Yang and Jingren Zhou and Junyang Lin and Kai Dang and Keming Lu and Keqin Bao and Kexin Yang and Le Yu and Mei Li and Mingfeng Xue and Pei Zhang and Qin Zhu and Rui Men and Runji Lin and Tianhao Li and Tianyi Tang and Tingyu Xia and Xingzhang Ren and Xuancheng Ren and Yang Fan and Yang Su and Yichang Zhang and Yu Wan and Yuqiong Liu and Zeyu Cui and Zhenru Zhang and Zihan Qiu},
      year={2025},
      eprint={2412.15115},
      archivePrefix={arXiv},
      primaryClass={cs.CL},
      url={https://arxiv.org/abs/2412.15115}, 
}

@inproceedings{clark2019boolq,
  title={Boolq: Exploring the surprising difficulty of natural yes/no questions},
  author={Clark, Christopher and Lee, Kenton and Chang, Ming-Wei and Kwiatkowski, Tom and Collins, Michael and Toutanova, Kristina},
  booktitle={Proceedings of the 2019 conference of the north American chapter of the association for computational linguistics: Human language technologies, volume 1 (long and short papers)},
  pages={2924--2936},
  year={2019}
}

@InProceedings{dagan2005pascal,
  title={The pascal recognising textual entailment challenge},
  author={Dagan, Ido and Glickman, Oren and Magnini, Bernardo},
  booktitle={Machine learning challenges workshop},
  pages={177--190},
  year={2005},
  organization={Springer}
}

@inproceedings{zellers2019hellaswag,
  title={Hellaswag: Can a machine really finish your sentence?},
  author={Zellers, Rowan and Holtzman, Ari and Bisk, Yonatan and Farhadi, Ali and Choi, Yejin},
  booktitle={Proceedings of the 57th annual meeting of the association for computational linguistics},
  pages={4791--4800},
  year={2019}
}

@inproceedings{sakaguchi2021winogrande,
  title={Winogrande: An adversarial winograd schema challenge at scale},
  author={Sakaguchi, Keisuke and Bras, Ronan Le and Bhagavatula, Chandra and Choi, Yejin},
  journal={Communications of the ACM},
  volume={64},
  number={9},
  pages={99--106},
  year={2021},
  publisher={ACM New York, NY, USA}
}

@article{clark2018arc,
  title={Think you have solved question answering? try arc, the ai2 reasoning challenge},
  author={Clark, Peter and Cowhey, Isaac and Etzioni, Oren and Khot, Tushar and Sabharwal, Ashish and Schoenick, Carissa and Tafjord, Oyvind},
  journal={arXiv preprint arXiv:1803.05457},
  year={2018}
}

@inproceedings{mihaylov2018openbookqa,
  title={Can a suit of armor conduct electricity? a new dataset for open book question answering},
  author={Mihaylov, Todor and Clark, Peter and Khot, Tushar and Sabharwal, Ashish},
  booktitle={Proceedings of the 2018 conference on empirical methods in natural language processing},
  pages={2381--2391},
  year={2018}
}

@misc{eval-harness,
  author       = {Lintang Sutawika and
                  Hailey Schoelkopf and
                  Leo Gao and
                  Baber Abbasi and
                  Stella Biderman and
                  Jonathan Tow and
                  ben fattori and
                  Charles Lovering and
                  farzanehnakhaee70 and
                  Jason Phang and
                  Anish Thite and
                  Fazz and
                  Thomas Wang and
                  Niklas Muennighoff and
                  Aflah and
                  sdtblck and
                  nopperl and
                  gakada and
                  tttyuntian and
                  researcher2 and
                  Chris and
                  Julen Etxaniz and
                  Hanwool Albert Lee and
                  Zdeněk Kasner and
                  Khalid and
                  Jeffrey Hsu and
                  Anjor Kanekar and
                  Pawan Sasanka Ammanamanchi and
                  Vicki Boykis and
                  AndyZwei},
  title        = {EleutherAI/lm-evaluation-harness: v0.4.2},
  month        = mar,
  year         = 2024,
  publisher    = {Zenodo},
  version      = {v0.4.2},
  doi          = {10.5281/zenodo.10829972},
  url          = {https://doi.org/10.5281/zenodo.10829972},
}

@inproceedings{han2015learning,
  title={Learning both weights and connections for efficient neural network},
  author={Han, Song and Pool, Jeff and Tran, John and Dally, William},
  journal={Advances in neural information processing systems},
  volume={28},
  year={2015}
}

@inproceedings{loshchilov2017adamw,
  title={Decoupled weight decay regularization},
  author={Loshchilov, Ilya and Hutter, Frank},
  journal={arXiv preprint arXiv:1711.05101},
  year={2017}

}

@article{merity2016wikitext2,
  title={Pointer sentinel mixture models},
  author={Merity, Stephen and Xiong, Caiming and Bradbury, James and Socher, Richard},
  journal={arXiv preprint arXiv:1609.07843},
  year={2016}
}

@inproceedings{devlin2019bert,
  title={Bert: Pre-training of deep bidirectional transformers for language understanding},
  author={Devlin, Jacob and Chang, Ming-Wei and Lee, Kenton and Toutanova, Kristina},
  booktitle={Proceedings of the 2019 conference of the North American chapter of the association for computational linguistics: human language technologies, volume 1 (long and short papers)},
  pages={4171--4186},
  year={2019}
}

@inproceedings{conneau2018xnli,
  title={XNLI: Evaluating cross-lingual sentence representations},
  author={Conneau, Alexis and Rinott, Ruty and Lample, Guillaume and Williams, Adina and Bowman, Samuel and Schwenk, Holger and Stoyanov, Veselin},
  booktitle={Proceedings of the 2018 conference on empirical methods in natural language processing},
  pages={2475--2485},
  year={2018}
}

@article{zhou2023ifeval,
  title={Instruction-following evaluation for large language models},
  author={Zhou, Jeffrey and Lu, Tianjian and Mishra, Swaroop and Brahma, Siddhartha and Basu, Sujoy and Luan, Yi and Zhou, Denny and Hou, Le},
  journal={arXiv preprint arXiv:2311.07911},
  year={2023}
}

@misc{hendrycks2021measuringmassivemultitasklanguage,
      title={Measuring Massive Multitask Language Understanding}, 
      author={Dan Hendrycks and Collin Burns and Steven Basart and Andy Zou and Mantas Mazeika and Dawn Song and Jacob Steinhardt},
      year={2021},
      eprint={2009.03300},
      archivePrefix={arXiv},
      primaryClass={cs.CY},
      url={https://arxiv.org/abs/2009.03300}, 
}

@misc{yang2025qwen3technicalreport,
      title={Qwen3 Technical Report}, 
      author={An Yang and Anfeng Li and Baosong Yang and Beichen Zhang and Binyuan Hui and Bo Zheng and Bowen Yu and Chang Gao and Chengen Huang and Chenxu Lv and Chujie Zheng and Dayiheng Liu and Fan Zhou and Fei Huang and Feng Hu and Hao Ge and Haoran Wei and Huan Lin and Jialong Tang and Jian Yang and Jianhong Tu and Jianwei Zhang and Jianxin Yang and Jiaxi Yang and Jing Zhou and Jingren Zhou and Junyang Lin and Kai Dang and Keqin Bao and Kexin Yang and Le Yu and Lianghao Deng and Mei Li and Mingfeng Xue and Mingze Li and Pei Zhang and Peng Wang and Qin Zhu and Rui Men and Ruize Gao and Shixuan Liu and Shuang Luo and Tianhao Li and Tianyi Tang and Wenbiao Yin and Xingzhang Ren and Xinyu Wang and Xinyu Zhang and Xuancheng Ren and Yang Fan and Yang Su and Yichang Zhang and Yinger Zhang and Yu Wan and Yuqiong Liu and Zekun Wang and Zeyu Cui and Zhenru Zhang and Zhipeng Zhou and Zihan Qiu},
      year={2025},
      eprint={2505.09388},
      archivePrefix={arXiv},
      primaryClass={cs.CL},
      url={https://arxiv.org/abs/2505.09388}, 
}
\appendix
\newpage
\section{Experimental Details} \label{supp:experimental_details}
This section provides comprehensive details on our experimental setup, including dataset descriptions, prompt/evaluation policies, implementation specifics, and reproducibility aspects.

\subsection{Datasets}\label{supp:dataset}
We evaluate on seven English benchmarks that are standard in prior pruning/efficiency work: BoolQ, RTE, HellaSwag, WinoGrande, ARC-Easy, ARC-Challenge, and OpenBookQA. Additionally, to verify the structural robustness and generalization of our method across multilingual and instruction-following domains, we extend our evaluation to XNLI, IFEval, and MMLU benchmarks in the additional analyses (Appendix~\ref{supp:analyses}). The comprehensive specifications for our evaluation pipeline, including the prompt wrapping policies and activation-collection input formats, are detailed in Table~\ref{tab:prompt-policy} and Table~\ref{tab:activation_input_templates}, respectively. For the seven source benchmarks, activation statistics are collected from their training splits. XNLI, IFEval, and MMLU are used only for post-pruning evaluation and do not contribute to mask construction.

\paragraph{BoolQ}
BoolQ \citep{clark2019boolq} is a yes/no reading comprehension dataset built from naturally occurring questions paired with supporting passages, primarily sourced from the web (e.g., Wikipedia). Each example contains a short question and a paragraph that may contain sufficient evidence to answer it. In our pipeline, we use the \textbf{train} split to collect activation statistics and the lm-eval-harness default configuration for zero-shot evaluation. During activation collection, we do not employ an explicit prompt; instead, inputs are formed by concatenating the passage and question. For evaluation, we rely on lm-eval’s standard yes/no template and report accuracy (and acc\_norm where applicable). Tokenization uses \texttt{max\_length=512} with padding/truncation; padding tokens are masked in our forward-hook aggregation.
\begin{table}[h]
\centering
\small
\renewcommand{\arraystretch}{1.2}
\begin{tabular}{lp{5.2cm}}
\toprule
\textbf{Component} & \textbf{Content} \\
\midrule
Question & do iran and afghanistan speak the same language \\
\midrule
Context & Persian language -- Persian, also known by its endonym Farsi, is one of the Western Iranian languages... It is primarily spoken in Iran, Afghanistan (officially known as Dari since 1958), and Tajikistan... \\
\midrule
Ground Truth & Yes \\
\bottomrule
\end{tabular}
\caption{Example data from \textsc{BoolQ}.}
\label{tab:dataset_boolq}
\end{table}

\paragraph{RTE}
RTE \citep{dagan2005pascal} evaluates two-way natural language inference over (premise, hypothesis) pairs. Labels are \emph{entailment} vs.\ \emph{not entailment}. We collect activations on the \textbf{train} split by concatenating the premise and hypothesis without a handcrafted prompt, then evaluate zero-shot with lm-eval-harness’ default NLI formatting and report accuracy. As with other tasks, we tokenize with \texttt{max\_length=512} and mask padding during aggregation to avoid biasing per-dimension importance estimates.
\begin{table}[h]
\centering
\small
\renewcommand{\arraystretch}{1.2}
\begin{tabular}{lp{5.2cm}}
\toprule
\textbf{Component} & \textbf{Content} \\
\midrule
Premise & No Weapons of Mass Destruction Found in Iraq Yet. \\
\midrule
Hypothesis & Weapons of Mass Destruction Found in Iraq. \\
\midrule
Ground Truth & not entailment \\
\bottomrule
\end{tabular}
\caption{Example data from \textsc{RTE}.}
\label{tab:dataset_rte}
\end{table}

\paragraph{HellaSwag}
HellaSwag \citep{zellers2019hellaswag} is a commonsense sentence completion benchmark where a short context must be continued with the most plausible of four candidate endings. It targets subtle lexical and pragmatic cues and is known to expose annotation artifacts if option order or formatting drifts from the original. In our setup, activation statistics are computed on the \textbf{train} split by concatenating the context with all four endings (no explicit instruction). Final accuracy is measured with lm-eval-harness zero-shot defaults that present the context and choices and parse answers as A–D. We keep the original option order and preserve special characters; long inputs are truncated at \texttt{512} tokens.
\begin{table}[h]
\centering
\small
\renewcommand{\arraystretch}{1.2}
\begin{tabular}{lp{5.2cm}}
\toprule
\textbf{Component} & \textbf{Content} \\
\midrule
Context & Then, the man writes over the snow covering the window of a car, and a woman wearing winter clothes smiles. then \\
\midrule
Options & A. , the man adds wax to the windshield and cuts it. \newline
          B. , a person board a ski lift, while two men supporting the head of the person wearing winter clothes snow as the we girls sled. \newline
          C. , the man puts on a christmas coat, knitted with netting. \newline
          D. , the man continues removing the snow on his car. \\
\midrule
Ground Truth & D \\
\bottomrule
\end{tabular}
\caption{Example data from \textsc{HellaSwag}.}
\label{tab:dataset_hellaswag}
\end{table}

\paragraph{WinoGrande}
WinoGrande \citep{sakaguchi2021winogrande} is a large-scale pronoun/coreference commonsense benchmark cast as a cloze task: a sentence with a single blank must be filled with one of two candidate nouns. We gather activations on the \textbf{train} split using only the raw sentence (without a formatted instruction or candidate list) to reflect content-driven activations; zero-shot evaluation then follows lm-eval’s default cloze template that shows the candidates and expects an answer in \{1,2\}. We report accuracy and escape underscores in LaTeX where necessary; truncation follows the same \texttt{512}-token policy.
\begin{table}[h]
\centering
\small
\renewcommand{\arraystretch}{1.2}
\begin{tabular}{lp{5.2cm}}
\toprule
\textbf{Component} & \textbf{Content} \\
\midrule
Sentence & Ian volunteered to eat Dennis's menudo after already having a bowl because \_\_\_ despised eating intestine. \\
\midrule
Options & 1. Ian \quad 2. Dennis \\
\midrule
Ground Truth & 2 (\textit{Dennis}) \\
\bottomrule
\end{tabular}
\caption{Example data from \textsc{WinoGrande}.}
\label{tab:dataset_winogrande}
\end{table}

\paragraph{ARC-Easy}
ARC-Easy \citep{clark2018arc} contains elementary-level science multiple-choice questions with four options. Items often involve factual recall plus light reasoning and unit/term fidelity. For activation collection, we use the \textbf{train} split and build inputs by concatenating the stem with the four option texts (no explicit prompt). Final zero-shot evaluation uses lm-eval-harness’ standard MCQ formatting, preserving the original option order, and we report accuracy (and acc\_norm where available). Scientific notation and symbols are kept verbatim; overly long stems are truncated after tokenization.
\begin{table}[h]
\centering
\small
\renewcommand{\arraystretch}{1.2}
\begin{tabular}{lp{5.2cm}}
\toprule
\textbf{Component} & \textbf{Content} \\
\midrule
Question & Which factor will most likely cause a person to develop a fever? \\
\midrule
Options & A. a leg muscle relaxing after exercise \newline
          B. a bacterial population in the bloodstream \newline
          C. several viral particles on the skin \newline
          D. carbohydrates being digested in the stomach \\
\midrule
Ground Truth & B \\
\bottomrule
\end{tabular}
\caption{Example data from \textsc{ARC-Easy}.}
\label{tab:dataset_arc_easy}
\end{table}

\paragraph{ARC-Challenge}
ARC-Challenge \citep{clark2018arc} is the harder subset of ARC that requires multi-hop reasoning or background knowledge beyond straightforward recall. We follow the same protocol as ARC-Easy: activations are collected on the \textbf{train} split using concatenated stem+choices without an explicit instruction, and evaluation is performed zero-shot with lm-eval defaults and original option order. As is typical, accuracy on ARC-Challenge is substantially lower than ARC-Easy for the same base model; our pruning analysis therefore reports both to illustrate difficulty sensitivity.
\begin{table}[h]           
\centering
\small
\renewcommand{\arraystretch}{1.2}
\begin{tabular}{lp{5.2cm}}  
\toprule
\textbf{Component} & \textbf{Content} \\
\midrule
Question & George wants to warm his hands quickly by rubbing them. Which skin surface will produce the most heat? \\
\midrule
Options & A. dry palms \quad B. wet palms \newline
C. palms covered with oil \newline
D. palms covered with lotion \\
\midrule
Ground Truth & A \\
\bottomrule
\end{tabular}
\caption{Example data from \textsc{ARC-Challenge}.}
\label{tab:dataset_arc_challenge}
\end{table}

\paragraph{OpenBookQA}
OpenBookQA \citep{mihaylov2018openbookqa} is a four-choice science QA dataset designed around a small “open book” of core facts. Many items benefit from combining the provided stem with general science knowledge. In our experiments we do \emph{not} augment with retrieval; activations on the \textbf{train} split are computed by concatenating the question stem and choice texts (no explicit prompt), and final zero-shot accuracy is obtained with lm-eval’s default MCQ template. This isolates pruning effects from external knowledge pipelines and keeps evaluation comparable to prior efficiency work.
\begin{table}[h]
\centering
\small
\renewcommand{\arraystretch}{1.2}
\begin{tabular}{lp{5.2cm}}
\toprule
\textbf{Component} & \textbf{Content} \\
\midrule
Question & The sun is responsible for \\
\midrule
Options & A. puppies learning new tricks \newline
          B. children growing up and getting old \newline
          C. flowers wilting in a vase \newline
          D. plants sprouting, blooming and wilting \\
\midrule
Ground Truth & D \\
\bottomrule
\end{tabular}
\caption{Example data from \textsc{OpenBookQA}.}
\label{tab:dataset_openbookqa}
\end{table}

\paragraph{XNLI}
XNLI \citep{conneau2018xnli} is a multilingual natural language inference benchmark constructed by extending the English MNLI dataset to multiple languages via human translation. Each example consists of a premise–hypothesis pair labeled as entailment, contradiction, or neutral, and is commonly used to evaluate cross-lingual sentence representations. In our experiments, we evaluate zero-shot performance without any language-specific adaptation or fine-tuning. Model predictions are obtained using lm-eval’s default NLI prompt template, and accuracy is reported by averaging results across languages. This setting allows us to assess how well pruning preserves distributed multilingual knowledge under a unified evaluation protocol.
\begin{table}[h]
\centering
\small
\renewcommand{\arraystretch}{1.2}
\begin{tabular}{lp{5.2cm}}
\toprule
\textbf{Component} & \textbf{Content} \\
\midrule
Premise & Yes, I sang for him and beautiful spirituals he liked. \\
\midrule
Hypothesis & I refused to sing for him. \\
\midrule
Options & A. Entailment \newline
          B. Contradiction \newline
          C. Neutral \\
\midrule
Ground Truth & B \\
\bottomrule
\end{tabular}
\caption{Example data from \textsc{XNLI}.}
\label{tab:dataset_xnli}
\end{table}

\paragraph{IFEval}
IFEval \citep{zhou2023ifeval} is a benchmark designed to evaluate instruction-following capability of large language models through verifiable constraints. Each instance specifies explicit requirements on the generated output, such as format, length, or content inclusion, which are automatically checked using rule-based evaluators rather than reference matching. In our evaluation, we follow the standard zero-shot setup without additional prompting or alignment techniques. Model outputs are assessed using both strict and loose accuracy at the prompt and instruction levels, following the benchmark's official evaluation. This evaluation focuses on whether pruning preserves the model’s ability to reliably follow instructions, independent of semantic correctness or stylistic quality.

\paragraph{MMLU}
MMLU \citep{hendrycks2021measuringmassivemultitasklanguage} is a broad knowledge and reasoning benchmark consisting of multiple-choice questions across 57 subjects spanning the humanities, social sciences, STEM, and
other academic and professional domains. Unlike the seven source
benchmarks used to construct the global pruning mask, MMLU is used
only as an additional post-pruning evaluation benchmark. We evaluate
the models under the zero-shot setting using
\texttt{lm-evaluation-harness} and report accuracy over four major
categories---Humanities, Other, Social Sciences, and STEM---as well
as the overall average. This evaluation assesses whether the global
mask transfers beyond the source-task suite while preserving broad
knowledge and reasoning capabilities.

\begin{table}[h]
\centering
\small
\renewcommand{\arraystretch}{1.2}
\begin{tabular}{lp{5.2cm}}
\toprule
\textbf{Component} & \textbf{Content} \\
\midrule
Prompt & Write a short essay about climate change. Your response must have exactly 3 paragraphs. Do not use the word "global" anywhere in the text. \\
\midrule
Constraints & 1. Number of paragraphs must be exactly 3. \newline
              2. Negative constraint: Forbidden word "global". \\
\midrule
Ground Truth & Verified automatically via rule-based software program execution. \\
\bottomrule
\end{tabular}
\caption{Example data from \textsc{IFEval}.}
\label{tab:dataset_ifeval}
\end{table}
\begin{table}[h]
\centering
\small
\renewcommand{\arraystretch}{1.2}
\begin{tabular}{lp{5.2cm}}
\toprule
\textbf{Component} & \textbf{Content} \\
\midrule
Subject & \texttt{high\_school\_biology} \\
\midrule
Question & In a population of giraffes, an environmental change occurs that favors individuals that are tallest. As a result, more of the taller individuals are able to obtain nutrients and survive to pass along their genetic information. This is an example of \\
\midrule
Options & A. directional selection. \newline
          B. stabilizing selection. \newline
          C. sexual selection. \newline
          D. disruptive selection. \\
\midrule
Ground Truth & A \\
\bottomrule
\end{tabular}
\caption{Example data from \textsc{MMLU}.}
\label{tab:dataset_mmlu}
\end{table}
\begin{table*}[!htbp]
\centering
\small
\renewcommand{\arraystretch}{1.2}
\begin{tabular}{lp{10.8cm}}
\toprule
\textbf{Stage} & \textbf{Basic Prompt / Source} \\
\midrule
Activation Collection (Pruning)
& No task instruction is added. Inputs are constructed from dataset
fields with lightweight field delimiters where applicable. \\
\midrule
Final Evaluation (lm-eval) &
\textbf{lm-eval-harness default prompts} with zero-shot (\texttt{--num\_fewshot 0}). No custom instruction added. \\
\bottomrule
\end{tabular}
\caption{Prompt policy used in our pipeline. For pruning analysis we do not use explicit templates; for final evaluation we rely on lm-eval-harness defaults.}
\label{tab:prompt-policy}
\end{table*}
\begin{table*}[!htbp]
\centering
\small
\renewcommand{\arraystretch}{1.2}
\begin{tabular}{llp{9.5cm}}
\toprule
\textbf{Dataset} & \textbf{Split} & \textbf{Input Format for Activation Collection} \\
\midrule
BoolQ & train & \texttt{\{passage\} Question: \{question\} Answer:} \\
RTE & train & \texttt{\{premise\} Question: \{hypothesis\} True or False?} \\
HellaSwag & train & \texttt{\{context\} \{choice\}} \\
WinoGrande & train & \texttt{\{sentence\}} \\
ARC-Easy & train & \texttt{Question: \{question\} Answer: \{choice\}} \\
ARC-Challenge & train & \texttt{Question: \{question\} Answer: \{choice\}} \\
OpenBookQA & train & \texttt{Question: \{question\} Answer: \{choice\}} \\
\bottomrule
\end{tabular}
\caption{Input formatting templates used for aggregating activation statistics across different benchmarks.}
\label{tab:activation_input_templates}
\end{table*}

\subsection{Measurement details}\label{supp:measurement}
We report \textbf{accuracy} (and \textbf{acc\_norm} where available) from \texttt{lm-eval-harness} with zero-shot (\texttt{--num\_fewshot 0}). 
All results use the official task loaders and default post-processing of lm-eval to ensure comparability.

For each benchmark, peak GPU memory is measured as the maximum
\texttt{torch.cuda.max\_memory\_allocated()} value across batches,
and we report the macro-average over the seven benchmarks.
Inference FLOPs are theoretically estimated as
$\mathrm{FLOPs}_{\mathrm{infer}}\approx 2NT$, where $N$ denotes
the number of unique parameters after accounting for weight tying
and $T$ denotes the average number of non-padding tokens per batch.
All measurements use forward inference with a batch size of 10 and
a maximum sequence length of 512.

\subsection{Baselines}\label{supp:baselines}
As our main baseline, we evaluate the unpruned base model with the same lm-eval settings. 
No custom prompt is added (zero-shot, defaults). 
For completeness, the basic evaluation templates (rendered by lm-eval) are summarized in Table~\ref{tab:basic-prompt-ours}.

\subsection{Implementation details}\label{supp:details}
We use Hugging Face \texttt{AutoModelForCausalLM}\slash\texttt{AutoTokenizer} with local checkpoints. For activation collection, we construct inputs from dataset fields, attach forward hooks to each MLP down-projection, and compute mean absolute activations over non-padded tokens (via the attention mask) to score hidden dimensions. Per task, dimensions are ranked and the lowest (r\%) are selected; a simple vote across tasks yields a consensus set, which we prune by zero-masking channels in linear layers whose input/output size matches the hidden width. Final accuracy is reported with \texttt{lm\_eval} (hf backend, zero-shot). All runs use a fixed seed (42) for reproducibility.

\begin{table*}[!htbp]
\centering
\small
\renewcommand{\arraystretch}{1.2}
\begin{tabular}{lp{11cm}}
\toprule
\textbf{Benchmark} & \textbf{Basic Prompt} \\
\midrule
BoolQ &
  Answer the question with \textbf{Yes or No} based on the context. \newline
  Q: \{question\} \newline
  Context: \{passage\} \newline
  Final format: \texttt{Answer: [Yes or No]} \\
\midrule
RTE &
  Determine whether the hypothesis is entailed by the premise. \newline
  Premise: \{premise\} \newline
  Hypothesis: \{hypothesis\} \newline
  Final format: \texttt{Answer: [entailment or not entailment]} \\
\midrule
HellaSwag &
  Choose the most plausible ending. \newline
  Context: \{ctx\} \newline
  Choices: (A) \{A\} (B) \{B\} (C) \{C\} (D) \{D\} \newline
  Final format: \texttt{Answer: [A/B/C/D]} \\
\midrule
WinoGrande &
  Fill the blank in the sentence with one candidate. \newline
  Sentence: \{sentence with \_\_\_ blank\} \newline
  Candidates: \{option1\}, \{option2\} \newline
  Final format: \texttt{Answer: [1 or 2]} \\
\midrule
ARC-Easy / ARC-Challenge &
  Select one correct option. \newline
  Question: \{question\} \newline
  Choices: (A) \{A\} (B) \{B\} (C) \{C\} (D) \{D\} \newline
  Final format: \texttt{Answer: [A/B/C/D]} \\
\midrule
OpenBookQA &
  Use basic science knowledge to answer. \newline
  Question: \{question\_stem\} \newline
  Choices: (A) \{A\} (B) \{B\} (C) \{C\} (D) \{D\} \newline
  Final format: \texttt{Answer: [A/B/C/D]} \\
\midrule
XNLI &
  Determine whether the hypothesis is entailed by the premise. \newline
  Premise: \{premise\} \newline
  Hypothesis: \{hypothesis\} \newline
  Final format: \texttt{Answer: [entailment, contradiction, or neutral]} \\
\midrule
IFEval &
  \{prompt\} \newline
  Final format: Generated free-form text output subject to strict
  rule-based constraint evaluation. \\
\midrule
MMLU &
  The following are multiple-choice questions about \{subject\}. \newline
  Question: \{question\} \newline
  Choices: (A) \{A\} (B) \{B\} (C) \{C\} (D) \{D\} \newline
  Final format: \texttt{Answer: [A/B/C/D]} \\
\bottomrule
\end{tabular}
\caption{Basic prompts used for our \textbf{evaluation}
(lm-eval-harness defaults).}
\label{tab:basic-prompt-ours}
\end{table*}

\newpage


\begin{table*}[t]
\caption{Zero-shot accuracy comparisons of \name{} against baseline pruning algorithms on Gemma-2 2B on seven NLP benchmarks. The best performances are marked in \textbf{bold}, and the runner-up is marked with \underline{underline}. Tasks with * are presented with \textit{acc\_norm} and remaining tasks with \textit{acc}.}
\label{tab:main_tab_2b_30}
\centering
\resizebox{0.95\textwidth}{!}{%
\begin{tabular}{@{}lccccccccc@{}} 
\toprule
\multirow{2}{*}{Method} & \multirow{2}{*}{\begin{tabular}[c]{@{}c@{}}Sparsity\end{tabular}} &
\multicolumn{7}{c}{Per-task Accuracies (\%)} &
\multirow{2}{*}{\begin{tabular}[c]{@{}l@{}}Average\\ Acc. (\%)\end{tabular}} \\ \cmidrule(lr){3-9}
 &  & \multicolumn{1}{l}{BoolQ} & RTE & HellaSwag* & WinoGrande & Arc-E* & Arc-C* & OBQA* & \\ \midrule

No pruning (Original) & 0 & 73.5 & 61.4 & 73.0 & 68.4 & 80.2 & 49.7 & 41.8 & 64.0 \\ \midrule



Magnitude - Dim & 30\% & 37.8 & 50.5 & 28.7 & 49.3 & 34.5 & 20.7 & 25.6 & 35.3 \\
SliceGPT                   & 30\% & 37.8 & 52.7 & 32.9 & 57.0 & 36.0 & 24.0 & 27.2 & \textbf{38.2} \\
PuDDing                    & 30\% & 39.6 & 46.2 & 26.3 & 49.6 & 28.6 & 24.5 & 27.2 & 34.6 \\
\midrule
DIET (Ours)                & 30\% & 41.3 & 49.1 & 28.8 & 49.3 & 35.9 & 21.5 & 24.2 & \underline{35.7} \\

\bottomrule
\end{tabular}%
}
\end{table*}


\begin{table*}[t]
\caption{Zero-shot accuracy comparisons of \name{} against baseline pruning algorithms on Gemma-2 9B on seven NLP benchmarks. The best performances are marked in \textbf{bold}, and the runner-up is marked with \underline{underline}. Tasks with * are presented with \textit{acc\_norm} and remaining tasks with \textit{acc}. All values are reported as percentages (\%).}
\label{tab:main_tab_9b_30}
\centering
\resizebox{0.95\textwidth}{!}{%
\begin{tabular}{@{}lccccccccc@{}} 
\toprule
\multirow{2}{*}{Method} & \multirow{2}{*}{\begin{tabular}[c]{@{}c@{}}Sparsity\end{tabular}} &
\multicolumn{7}{c}{Per-task Accuracies (\%)} &
\multirow{2}{*}{\begin{tabular}[c]{@{}l@{}}Average\\ Acc. (\%)\end{tabular}} \\ \cmidrule(lr){3-9}
 &  & \multicolumn{1}{l}{BoolQ} & RTE & HellaSwag* & WinoGrande & Arc-E* & Arc-C* & OBQA* & \\ \midrule

No pruning (Original) & 0 & 84.4 & 67.9 & 80.0 & 74.2 & 87.8 & 66.1 & 47.2 & 72.5 \\ \midrule



Magnitude - Dim & 30\% & 53.9 & 52.7 & 32.6 &	51.9 & 44.1 & 22.6 &	25.0 & 40.4 \\
SliceGPT                   & 30\% & 37.9 & 53.8 & 40.8 & 63.2 &	46.2 & 27.2 & 31.6 & \underline{43.0} \\
PuDDing                    & 30\% & 50.5 & 66.1 & 49.2 & 60.0 & 52.1 & 32.2 & 31.6 & \textbf{48.8} \\
\midrule
DIET (Ours)                & 30\% & 61.2 & 53.1 & 32.3 & 52.9 &	43.1	& 22.5 &	24.2 &	41.3 \\

\bottomrule
\end{tabular}%
}
\end{table*}

\section{Additional Analyses}\label{supp:analyses}


\paragraph{Performance under aggressive sparsification.}
To explore the limits of compression, we extend our evaluation to an aggressive 30\% sparsity regime, with results summarized in Table~\ref{tab:main_tab_2b_30} and Table~\ref{tab:main_tab_9b_30}. At 30\% sparsity, all evaluated methods suffer substantial accuracy
degradation. Unlike the results at 10\% and 20\% sparsity, however,
\name{} does not consistently outperform the structured pruning
baselines in this aggressive regime. This result suggests that the benefit of task-wise dimension aggregation reduces under extreme model compression. The shared degradation across methods indicates that extreme sparsification is inherently challenging, although the relative ranking suggests that \name{} becomes more sensitive in this regime.

\paragraph{Robustness across diverse capabilities.}
To evaluate how well \name{} preserves broader linguistic and functional dimensions, we report additional zero-shot results on XNLI (multilingual understanding), IFEval (verifiable instruction-following), and MMLU (broad knowledge and reasoning) in Tables~\ref{tab:xnli_2b}, \ref{tab:xnli_9b}, \ref{tab:ifeval_all}, and \ref{tab:mmlu_category}. At 10\% sparsity, \name{} consistently outperforms weight-magnitude baselines on XNLI for both Gemma-2-2B and 9B backbones, showing a more pronounced margin in the smaller model scale. On IFEval, \name{} shows mixed performance across model scales and sparsity levels, with its clearest gains appearing on
Gemma-2 9B at 10\% sparsity. On MMLU, \name{} consistently outperforms Magnitude-Dim across all four broad categories at 20\% sparsity, indicating that the resulting global mask also preserves broad knowledge and reasoning capabilities beyond the original benchmark suite.
\begin{table*}[t]
\caption{Zero-shot accuracy comparisons of \name{} against baseline pruning algorithms on Gemma-2 2B on XNLI (15 languages). The best average performances at each sparsity level are marked in \textbf{bold}. All values are reported as percentages (\%).}
\label{tab:xnli_2b}
\centering
\resizebox{\textwidth}{!}{%
\begin{tabular}{@{}lcccccccccccccccccc@{}}
\toprule
\multirow{2}{*}{Method} & \multirow{2}{*}{Sparsity} &
\multicolumn{15}{c}{Per-language Accuracies (\%)} &
\multirow{2}{*}{\begin{tabular}[c]{@{}l@{}}Average\\ Acc. (\%)\end{tabular}} \\ \cmidrule(lr){3-17}
 &  & en & fr & es & de & el & bg & ru & tr & ar & vi & th & zh & hi & sw & ur & \\ \midrule

No pruning (Original) & 0 & 53.9 & 49.3 & 46.8 & 49.0 & 39.1 & 44.1 & 49.0 & 45.0 & 33.2 & 47.0 & 44.4 & 35.5 & 43.5 & 41.0 & 34.1 & 43.7 \\ \midrule

Magnitude-Dim & 10\% & 37.1 & 37.2 & 34.5 & 34.3 & 34.2 & 34.7 & 36.4 & 33.6 & 34.1 & 33.5 & 33.5 & 34.0 & 34.8 & 34.1 & 34.3 & 34.7 \\
DIET (Ours) & 10\% & 49.8 & 41.9 & 45.3 & 42.6 & 35.6 & 39.2 & 38.7 & 34.2 & 33.5 & 34.3 & 37.6 & 32.8 & 37.7 & 34.2 & 33.3 & \textbf{38.0} \\ \midrule

Magnitude-Dim & 20\% & 35.0 & 34.9 & 33.7 & 33.4 & 33.6 & 33.7 & 36.4 & 32.9 & 33.6 & 33.1 & 33.5 & 33.3 & 36.1 & 33.6 & 33.4 & 34.0 \\
DIET (Ours) & 20\% & 40.0 & 34.1 & 36.6 & 33.7 & 33.4 & 34.1 & 33.5 & 33.4 & 33.2 & 33.4 & 33.6 & 33.3 & 35.1 & 34.6 & 33.3 & \textbf{34.4} \\

\bottomrule
\end{tabular}%
}
\end{table*}
\begin{table*}[ht!]
\caption{Zero-shot accuracy comparisons of \name{} against baseline pruning algorithms on Gemma-2 9B on XNLI (15 languages). The best average performances at each sparsity level are marked in \textbf{bold}. All values are reported as percentages (\%).}
\label{tab:xnli_9b}
\centering
\resizebox{\textwidth}{!}{%
\begin{tabular}{@{}lcccccccccccccccccc@{}}
\toprule
\multirow{2}{*}{Method} & \multirow{2}{*}{Sparsity} &
\multicolumn{15}{c}{Per-language Accuracies (\%)} &
\multirow{2}{*}{\begin{tabular}[c]{@{}l@{}}Average\\ Acc. (\%)\end{tabular}} \\ \cmidrule(lr){3-17}
 &  & en & fr & es & de & el & bg & ru & tr & ar & vi & th & zh & hi & sw & ur & \\ \midrule

No pruning (Original) & 0 & 54.0 & 51.6 & 51.0 & 51.1 & 45.3 & 48.6 & 49.0 & 50.4 & 33.8 & 45.2 & 47.5 & 33.4 & 46.6 & 46.5 & 44.4 & 46.6 \\ \midrule

Magnitude-Dim & 10\% & 52.3 & 44.0 & 49.5 & 49.0 & 42.7 & 38.6 & 41.5 & 42.9 & 33.6 & 47.4 & 38.8 & 33.3 & 40.2 & 40.7 & 34.1 & 41.9 \\
DIET (Ours) & 10\% & 52.6 & 50.2 & 47.0 & 48.7 & 41.6 & 37.4 & 47.3 & 44.7 & 34.0 & 42.9 & 40.8 & 33.9 & 44.5 & 38.8 & 38.1 & \textbf{42.8} \\ \midrule

Magnitude-Dim & 20\% & 45.0 & 35.2 & 43.5 & 42.3 & 38.8 & 35.1 & 35.6 & 35.3 & 33.7 & 39.9 & 35.9 & 33.1 & 39.2 & 37.1 & 33.4 & \textbf{37.5} \\
DIET (Ours) & 20\% & 43.7 & 43.3 & 37.1 & 43.5 & 38.6 & 34.9 & 37.2 & 38.8 & 33.5 & 36.1 & 34.5 & 33.7 & 37.6 & 34.5 & 34.1 & 37.4 \\

\bottomrule
\end{tabular}%
}
\end{table*}
\begin{table*}[ht!]
\caption{Zero-shot accuracy comparisons on IFEval for Gemma-2 2B and 9B. All values are reported as accuracy scores.}
\label{tab:ifeval_all}
\centering
\resizebox{0.9\textwidth}{!}{%
\begin{tabular}{@{}lc|cccc|cccc@{}}
\toprule
\multirow{3}{*}{Method} & \multirow{3}{*}{Sparsity} & \multicolumn{4}{c|}{\textbf{Gemma-2 2B}} & \multicolumn{4}{c}{\textbf{Gemma-2 9B}} \\ \cmidrule(lr){3-6} \cmidrule(l){7-10}
 &  & \multicolumn{2}{c}{Strict Acc} & \multicolumn{2}{c|}{Loose Acc} & \multicolumn{2}{c}{Strict Acc} & \multicolumn{2}{c}{Loose Acc} \\ \cmidrule(lr){3-4} \cmidrule(lr){5-6} \cmidrule(lr){7-8} \cmidrule(l){9-10} 
 &  & Prompt & Inst & Prompt & Inst & Prompt & Inst & Prompt & Inst \\ \midrule

Original & 0 & 0.1368 & 0.2710 & 0.1386 & 0.2758 & 0.1497 & 0.2698 & 0.1590 & 0.2770 \\ \midrule

Magnitude-Dim & \multirow{2}{*}{10\%} & 0.1331 & 0.2698 & 0.1405 & 0.2758 & 0.1497 & 0.2614 & 0.1608 & 0.2710 \\
\name{} (Ours) &  & 0.1035 & 0.2158 & 0.1183 & 0.2278 & 0.1553 & 0.2842 & 0.1590 & 0.2902 \\ \midrule

Magnitude-Dim & \multirow{2}{*}{20\%} & 0.1054 & 0.2302 & 0.1054 & 0.2302 & 0.1165 & 0.2398 & 0.1220 & 0.2434 \\
\name{} (Ours) &  & 0.1091 & 0.2206 & 0.1109 & 0.2230 & 0.1146 & 0.2326 & 0.1146 & 0.2326 \\

\bottomrule
\end{tabular}%
}
\end{table*}
\begin{table*}[!t]
\caption{
Category-wise zero-shot MMLU accuracy (\%) on Gemma-2 2B
at 20\% structured pruning.
$\Delta$ denotes the absolute improvement of \name{} over
Magnitude-Dim in percentage points.
}
\label{tab:mmlu_category}
\centering
\small
\renewcommand{\arraystretch}{1.15}
\setlength{\tabcolsep}{10pt}

\begin{tabular*}{0.72\textwidth}{
    @{\extracolsep{\fill}}
    lccc
    @{}
}
\toprule
\textbf{Category}
& \textbf{Magnitude-Dim}
& \textbf{\name{} (Ours)}
& \textbf{$\Delta$ (pp)} \\
\midrule
Humanities
& 23.91
& \textbf{28.48}
& $+4.57$ \\

Other
& 27.71
& \textbf{32.57}
& $+4.86$ \\

Social Sciences
& 26.00
& \textbf{31.78}
& $+5.78$ \\

STEM
& 26.29
& \textbf{31.81}
& $+5.52$ \\
\midrule
Overall
& 25.74
& \textbf{30.86}
& $+5.12$ \\
\bottomrule
\end{tabular*}
\end{table*}

\paragraph{Global mask vs.\ per-task masks.}
We compare our merged global mask to models pruned with a single-task mask derived from \textit{arc\_easy} and \textit{arc\_challenge}. As shown in Table~\ref{tab:per_task}, the global mask improves average zero-shot accuracy over single-task mask applied models. The gains are particularly notable on \textit{HellaSwag}, where the global mask improves accuracy by 2.8 percentage points over both single-task masks, and on \textit{RTE}, where it yields improvements of 2.4 and 2.5 percentage points over the \textit{Arc-Easy} and \textit{Arc-Challenge} masks, respectively.
\begin{table*}[!htbp]
\caption{Zero-shot accuracy on Gemma-2 2B with masks derived from \textit{arc\_easy}, \textit{arc\_challenge}, and all tasks. Tasks with * are presented with \textit{acc\_norm} and remaining tasks with \textit{acc}.}
\label{tab:per_task}
\centering
\resizebox{0.95\textwidth}{!}{%
\begin{tabular}{@{}lccccccccc@{}}
\toprule
\multirow{2}{*}{Method} & \multirow{2}{*}{Sparsity} &
\multicolumn{7}{c}{Per-task Accuracies (\%)} &
\multirow{2}{*}{\begin{tabular}[c]{@{}l@{}}Average\\ Acc. (\%)\end{tabular}} \\
\cmidrule(lr){3-9}
 &  & BoolQ & RTE & HellaSwag* & WinoGrande & Arc-E* & Arc-C* & OBQA* & \\
\midrule

Arc-Easy Mask        & 20\% & 62.9 & 52.5 & 39.0 & 53.1 & 48.6 & 25.2 & 26.8 & 44.0 \\
Arc-Challenge Mask   & 20\% & 63.4 & 52.4 & 39.0 & 51.5 & 48.0 & 24.5 & 26.8 & 43.6 \\
\midrule
All Tasks (Ours)     & 20\% & 63.3 & 54.9 & 41.8 & 53.2 & 49.8 & 24.1 & 27.6 & 45.0 \\

\bottomrule
\end{tabular}%
}
\end{table*}

\paragraph{Continuous score merging vs.\ binary voting.}
Beyond binary selectors, we also merge \emph{continuous} task-wise importance scores: for each dimension, we construct a vector of normalized activation magnitudes across tasks, aggregate it to obtain a scalar score, and rank dimensions accordingly (smaller scores indicate less informative dimensions). We then select a global fraction of dimensions according to the target sparsity (20\%) based on this ranking and prune the model. As reported in Table~\ref{tab:continuous}, the pruned model under this continuous merging scheme attains accuracy comparable to the binary vote–based \name{}, showing that activation magnitudes provide sufficiently informative signals and that \name{} is robust to the merging granularity (binary vs.\ continuous).
\begin{table*}[!htbp]
\caption{Zero-shot accuracy comparisons of continuous score merging and our discrete (vote-based) mask on Gemma-2 2B at 20\% sparsity. Tasks with * are presented with \textit{acc\_norm} and remaining tasks with \textit{acc}.}
\label{tab:continuous}
\centering
\resizebox{0.95\textwidth}{!}{%
\begin{tabular}{@{}lccccccccc@{}}
\toprule
\multirow{2}{*}{Method} & \multirow{2}{*}{Sparsity} &
\multicolumn{7}{c}{Per-task Accuracies (\%)} &
\multirow{2}{*}{\begin{tabular}[c]{@{}c@{}}Average\\ Acc. (\%)\end{tabular}} \\
\cmidrule(lr){3-9}
 &  & BoolQ & RTE & HellaSwag* & WinoGrande & Arc-E* & Arc-C* & OBQA* &  \\
\midrule

Continuous     & 20\% & 63.3 & 52.7 & 42.1 & 52.6 & 50.4 & 24.5 & 26.8 & 44.6 \\

\midrule
Voting (Ours)  & 20\% & 63.3 & 54.9 & 41.8 & 53.2 & 49.8 & 24.1 & 27.6 & 45.0 \\

\bottomrule
\end{tabular}%
}
\end{table*}

\section{Usage of AI Tools}\label{supp:AI}
During manuscript preparation, we used AI-based language tools for editorial revisions such as grammar correction and readability improvements. These tools did not contribute to the conception of the study, methodological development, experimental design, data analysis, or the articulation of scientific conclusions. All substantive content originates from the authors; AI assistance was restricted to copy-editing and has no bearing on the work’s originality.



\end{document}